\documentclass{article}

\usepackage{microtype}
\usepackage{graphicx}
\usepackage{booktabs} 

\usepackage[colorlinks]{hyperref}

\usepackage[accepted]{icml2021}

\usepackage{url}
\usepackage{todonotes}
\usepackage{amsfonts}       
\usepackage{amsmath}
\usepackage{microtype}      
\usepackage{nicefrac}       
\usepackage{graphicx}
\usepackage{caption}
\usepackage{subcaption}
\usepackage{natbib}
\usepackage{layouts}
\usepackage[noabbrev,nameinlink,capitalize]{cleveref}

\begin{document}


\icmltitlerunning{Joint Attention for Multi-Agent Coordination and Social Learning}

\twocolumn[
\icmltitle{Joint Attention for Multi-Agent Coordination and Social Learning}



\icmlsetsymbol{equal}{*}

\begin{icmlauthorlist}
\icmlauthor{Dennis Lee}{goo}
\icmlauthor{Natasha Jaques}{goo}
\icmlauthor{Chase Kew}{goo}
\icmlauthor{Jiaxing Wu}{goo}
\icmlauthor{Douglas Eck}{goo}
\icmlauthor{Dale Schuurmans}{goo}
\icmlauthor{Aleksandra Faust}{goo}
\end{icmlauthorlist}

\icmlaffiliation{goo}{Google, Mountain View, USA}

\icmlcorrespondingauthor{Dennis Lee}{ldennis@google.com}
\icmlcorrespondingauthor{Natasha Jaques}{natashajaques@google.com}

\icmlkeywords{Machine Learning, ICML}

\vskip 0.3in
]



\printAffiliationsAndNotice{} 


\begin{abstract}
Joint attention---the ability to purposefully coordinate attention with another agent, and mutually attend to the same thing---is a critical component of human social cognition. In this paper, we ask whether joint attention can be useful as a mechanism for improving multi-agent coordination and social learning. We first develop deep reinforcement learning (RL) agents with a recurrent visual attention architecture. We then train agents to minimize the difference between the attention weights that they apply to the environment at each timestep, and the attention of other agents. Our results show that this joint attention incentive improves agents' ability to solve difficult coordination tasks, by reducing the exponential cost of exploring the joint multi-agent action space. Joint attention leads to higher performance than a competitive centralized critic baseline across multiple environments. Further, we show that joint attention 
enhances agents' ability to learn from experts present in their environment, even when completing hard exploration tasks that do not require coordination. Taken together, these findings suggest that joint attention may be a useful inductive bias for multi-agent learning.
\end{abstract}




         
\newcommand{\BibTeX}{\rm B\kern-.05em{\sc i\kern-.025em b}\kern-.08em\TeX}

\section{Introduction}
\textit{Joint attention} (JA) \cite{tomasello1995joint} is an important milestone in human cognitive development \cite{mundy2000eeg}. Sometimes referred to as `social attention coordination', JA is the ability to infer what another agent is attending to and purposefully coordinate attention with them, so that you are mutually attending to each other and the same object or event. JA plays a pivotal role in human social intelligence \cite{kaplan2006challenges,mundy1990longitudinal}, and is considered a precursor to understanding others' thoughts, beliefs, and intentions \cite{tomasello1995joint}. 


In this paper, we investigate whether a mechanism inspired by human joint attention can act as a useful inductive bias for multi-agent reinforcement learning (MARL). While training multiple agents to learn a coordinated policy would benefit a number of real-world applications, including robotics \cite{ota2006multi}, autonomous driving (e.g. \cite{shalev2016safe}) and sensor and communication networks (e.g. \cite{choi2009distributed}), it poses a difficult reinforcement learning (RL) problem. This is because the multi-agent joint action space increases exponentially with the number of agents, leading to a  combinatorial explosion that makes exploration in MARL a formidable challenge and impairs its ability to scale to many agents \cite{hernandez2019survey,zhang2019multi,claus1998dynamics}. Past work has addressed this problem through various ways to factor the joint value function (e.g. \cite{lowe2017multi,rashid2018qmix}), or through computationally expensive intrinsic motivations designed to achieve better coordination \cite{jaques2018intrinsic}.

Instead, we propose a simple incentive based on joint attention, in which agents are rewarded for matching their attention with other agents in the same environment. This can be considered an \textit{intrinsic motivation} \cite{chentanez2005intrinsically,schmidhuber2010formal} to maintain joint attention with other agents. To support this objective, we equip agents with a recurrent deep neural network architecture which uses \textit{visual attention} to selectively weight the most important elements of the environment state while ignoring the rest. 
We then give agents an additional reward based on how well their visual attention map matches that of the other agents. We hypothesize that focusing on the same elements of the environment at the same time will reduce the cost of multi-agent exploration. Although prior work has explored how joint attention can be applied to human robot interaction (e.g. \cite{huang2010joint}), to the best of our knowledge we are the first to propose joint attention as a way to facilitate multi-agent coordination. 

While the mechanism we study here is not directly equivalent to human joint attention, 
we believe that the neural network architecture we have developed shares important similarities with human attention. We use a softmax layer across a limited number of attention heads to create a bottleneck, forcing agents to selectively attend to only a few elements of the environment at once, while filtering out irrelevant details. Human selective attention also involves filtering irrelevant details to only attend to relevant stimuli \cite{tang2020neuroevolution}. Secondly, Tomasello---who proposed the concept of joint attention \cite{tomasello1995joint}---describes attention as ``goal-driven directed perception''. Our agents condition their attention on the contents of their recurrent policy network, which enables them to direct their attention based on their interaction history with the environment, in order to attain higher rewards. 

Our experiments demonstrate that the proposed joint attention mechanism enables agents to learn more quickly and obtain higher final reward in a series of multi-agent coordination tasks, exceeding the performance of competitive centralized-critic baselines \cite{lowe2017multi}. Further, we show that when novice agents are placed in an environment with pre-trained experts, joint attention allows them to learn the task more quickly by guiding them to attend to the most salient elements of the environment. This is true even for environments that do not require multi-agent coordination, suggesting joint attention aids social learning more broadly.
In humans, joint attention enables both social coordination and learning from an expert caregiver. In this paper, we have demonstrated that our mechanism, inspired by joint attention, can provide similar benefits to RL agents.

\section{Related work}
\textbf{Human-robot interaction.}
The importance of joint attention in social behavior has been widely recognized by the human robot interaction (HRI) community (e.g. \cite{huang2010joint,deak2001emergence,thomaz2005embodied,marin2009towards}). For example, \citet{huang2010joint} showed that when robots engage in joint attention with a human, humans develop a better model of the robot, can perform tasks with it more easily, and perceive it as more competent and socially interactive. Many HRI studies focus on embodied aspects of joint attention, such as following a human's gaze, or using pointing gestures, and often involve scripted procedures \cite{hoffman2006probabilistic,carlson2004computational,nagai2003constructive,kozima2001robot,imai2003physical,scassellati1998imitation}. However, this early work focuses on enhancing HRI, rather than multi-agent interaction. Further, it does not make use of the modern deep neural network notion of attention. 

\textbf{Multi-agent reinforcement learning.}
The exponential increase in the action space inherent in MARL makes learning a joint policy which simultaneously controls all agents prohibitively expensive, especially as the number of agents increases \cite{hernandez2019survey,zhang2019multi}. Therefore, many prior works have taken a Centralized Training Decentralized Execution (CTDE) approach, learning policies that can be executed independently but are trained together. Often, this takes the form of agents that share a joint value function. Examples of this approach include MADDPG \cite{lowe2017multi}, Value Decomposition Networks (VDN) \cite{sunehag2018value}, and QMIX \cite{rashid2018qmix}. COMA \cite{foerster2017counterfactual} uses counterfactual reasoning to enable each agent to isolate its impact on the joint value function. In contrast, our approach does not require training a centralized critic. 

\citet{jaques2018intrinsic} propose an intrinsic social motivation based on rewarding agents for increasing their influence over other agents' actions. This mechanism increased coordination for the social dilemmas studied in original paper, but could potentially incentivize non-cooperative behavior in other environments \cite{jaques2018intrinsic}. In addition, calculating the social influence reward leads to a quadratic increase in computational complexity. Our approach is much less computationally expensive, and is designed to more directly reward cooperative behavior.

Attention has emerged as a useful tool for multi-agent learning. For example, \citet{iqbal2019actor} use centralized critics with attention to select relevant information for each agent. Other papers have focused on using attention in graph networks as an efficient structure for MARL (e.g. \cite{tacchetti2018relational,liu2020multi,jiang2018graph,malysheva2019magnet,luo2019multi,das2019tarmac}). In spite of this, to the best of our knowledge the concept of joint attention has not been deployed as a method for improving multi-agent coordination. Although \citet{falkner2020learning} use the term ``joint attention'', they are actually referring to applying attention to the joint action space, and thus are not aligned with the psychological definition of joint attention proposed by \citet{tomasello1995joint} which we study in this paper. \citet{kobayashi2012cooperative} build a mechanism to compute the degree of ``eye contact'' between agents in a grid world, however they but do not build on the concept of joint attention, and compute eye contact using a fixed formula specific to grid worlds, rather than using neural network attention to enable agents to learn what to attend to in the environment. 

\begin{figure*}[ht!]
\includegraphics[width=\linewidth]{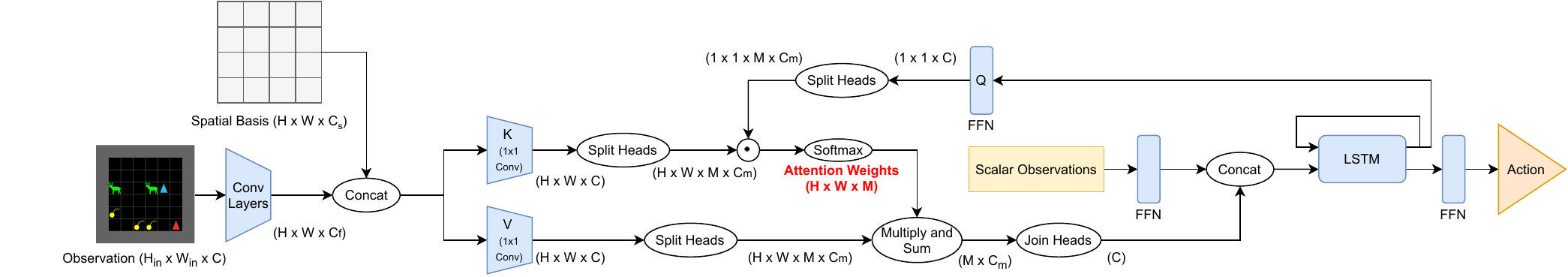} 
\caption{Recurrent visual attention architecture for each agent. Convolution (conv) layers process the input, after which a spatial basis matrix $S$ is appended to enable retention of location information. Attention is computed based on queries from the recurrent LSTM policy network, meaning that attention is a top-down, goal-directed selection process based on optimizing future rewards given the past history of interaction. Queries are applied to the keys matrix $K$, to produce attention weights $A$, which filtering the values image $V$. The result is input to the LSTM, which produces the next action. }
\label{fig:attention_architecture}
\end{figure*}

\textbf{Attention architectures.}
Since attention was first proposed as a mechanism for improving deep neural networks, it has been used with impressive success in a variety of architectures (e.g. \citet{bahdanau2014neural,vaswani2017attention}). Visual attention enables selectively attending to the important components of an image (e.g. \citet{mnih2014recurrent,woo2018cbam,bello2019attention} For the purposes of this paper, we are concerned with papers which incorporate visual attention into an RL policy (e.g. \citet{choi2017multi}). For example, \citet{tang2020neuroevolution} use a modified form of attention to determine the most important image patches for a recurrent policy network. Unlike our approach however, the state of the recurrent network is not used to inform the attention mechanism. Our architecture is more similar to that of \citet{mott2019towards}, which was developed to improve the interpretability of RL agents by learning attention maps that are readable by humans. 

\section{Background}
We focus on multi-agent reinforcement learning (MARL) environments defined by the tuple $\langle \mathcal{S}, \mathcal{A}, \mathcal{T}, \mathcal{R}, N \rangle$, where $N$ is the number of agents, and $s \in \mathcal{S}$ is the state of the environment. At each timestep $t$, each agent $k$ chooses a discrete action $a^k_t \in \mathcal{A}$. Agents act simultaneously and there is no notion of turn-taking. Let $\mathcal{A}^N$ be the joint action space, and $\vec{a}_t$ be the vector containing the actions of all agents for timestep $t$. The transition function depends on the joint action space: $\mathcal{T}: \mathcal{S} \times \mathcal{A}^N \times \mathcal{S} \rightarrow [0,1]$, as does the 
reward function $\mathcal{R}: \mathcal{S} \times \mathcal{A}^N \rightarrow \mathbb{R}^N$.
Each agent $k$ is attempting to maximize its own reward by learning a policy $\pi^k$ that optimizes the total expected discounted future reward: $J(\pi^k) = \mathbb{E}_\pi \Big[ \sum_{t=0}^{\infty} \gamma^t \, r^k_{t+1} \, | \, s_0 \Big]$, given a starting state $s_0$ and a discount factor $\gamma \in [0,1]$. Note that agents cannot directly observe other agents actions, states, or rewards, and do not share parameters. To simplify notation, when we discuss the architecture and learning objectives for a single agent we forego the superscript notation.


\section{Model architecture}
In order to train agents to sustain joint attention with other agents, we must first have a reliable method for estimating agents' visual attention. Therefore, we developed a novel deep network architecture for computing goal-directed visual attention for multi-agent RL (shown in Figure \ref{fig:attention_architecture}). Given an image of the environment state $s$ with height $h$ and width $w$, our goal is to apply visual attention to obtain an $h \times w$ Attention Map $A$ for each agent, indicating how strongly it is attending to each element of the environment. We can then calculate the similarity between agents' attention maps in order to compute the joint attention incentive. The code for our agents is available in open-source at $<$URL redacted$>$. 

Given an input image $X \in \mathbb{R}^{h \times w \times c}$ with $c$ channels, we first process it with a convolutional (conv) layer to extract a matrix of features, $F \in \mathbb{R}^{h \times w \times c_f}$, where $c_f$ is the number of conv filters. Inspired by \cite{mott2019towards}, we then append a spatial basis matrix, which is a fixed tensor $S \in \mathbb{R}^{h \times w \times c_s}$ to $F$. The basis provides spatial information, such that when the attention layer compresses the image, information about where the compressed information was located in the input is retained. Equations for computing $S$ can be found in Section \ref{sec:spatial_basis} of the Appendix. 

After appending $S$ to $F$, we input the resulting $h \times w \times c_f + c_s$ matrix into a multi-head attention layer similar to the kind employed by Transformers \cite{vaswani2017attention}. It uses two additional conv layers, which produce a matrix of keys, $K \in \mathbb{R}^{h \times w \times m \times c_m}$, and values $V \in \mathbb{R}^{h \times w \times m \times c_m}$, where $m$ is the number of attention heads, and $c_m$ is the number of features per head. The attention weights are computed by taking the inner product between a set query vectors $Q \in \mathbb{R}^{m \times c_m}$ and $K$. Specifically, for each query vector $q^i$ in $Q$, we obtain attention logits $\tilde{a}^i_{x,y}$:
\begin{align}
    \tilde{a}^i_{x,y} = \sum_j q^i_j K^i_{x,y,j}
\end{align}
where $i$ indexes the attention head, and $j$ indexes the feature within the query and key matrix (which have $c_m$ features). 

To compute the final normalized attention map $A$, we apply a softmax function to the logits for each attention head:
\begin{align}
    a^i_{x,y} = \frac{\exp(\tilde{a}^i_{x,y})}{\sum_{x',y'}\exp(\tilde{a}^i_{x',y'})}
    \label{eq:attention}
\end{align}
where $a^i_{x,y}$ is an entry in $A$. The softmax forces each head to concentrate attention on only a few elements of the image. 

We use the computed attention to filter the input image to selectively focus on certain elements, by performing an element-wise multiplication between $A$ and the values matrix $V$. Specifically, the filtered output matrix $O \in \mathbb{R}^{m \times c_m}$ produced by the visual attention portion of the network is computed according to $o^i_j = \sum_{x,y}a^i_{x,y}v_{x,y,j}$. The output $O$ is then fed into the rest of the agent's policy network.

We parameterize the policy using a recurrent Long Short Term Memory (LSTM) \cite{hochreiter1997long} recurrent neural network (RNN). 
Since the policies of other agents are not known, the environment is inherently non-Markov. Information about the other agents' past behavior is not contained in the state $s_t$, but is predictive of the next state, $s_{t+1}$. Therefore, we choose a recurrent architecture to enable agents to remember the history of states in which they have observed other agents, and improve their ability to model other agents' policies.
The LSTM takes both the output $O$, and the agent's position and direction, $p$, as input. 

Let $h_t = f_L(O_t, p_t, h_{t-1}; \theta_L)$ be the hidden cell contents of the LSTM for timestep $t$, where the LSTM is represented by function $f_L$, and $\theta_L$ are the LSTM parameters. The LSTM state is used to compute both the RL policy, as well as produce the query vectors for the attention layer. Specifically, the query vectors for timestep $t$ are computed using a feed-forward network $f_Q$, parameterized by $\theta_Q$, applied to the LSTM state at timestep $t-1$:
\begin{align}
    Q_t &= f_Q(f_L(O_{t-1}, h_{t-2}; \theta_L); \theta_Q) \\
        &= f_Q(h_{t-1}; \theta_Q)
\end{align}
This modeling choice means that the attention queries are produced top-down, based on the LSTM policy's objective of maximizing reward, and conditioned on the history of interaction with the environment and other agents. 

To produce the RL policy $\pi(a_t|s_t)$, we pass $h_t$ through a final feed-forward layer. We train the policy using Proximal Policy Gradients (PPO) \citep{schulman2017proximal}, as PPO has been found to be highly effective for multi-agent deep RL \cite{benchmarking}. More details about the training objective and architecture are given in Appendix \ref{sec:training}.


\section{Incentivizing joint attention}
Joint attention requires sustaining attention on the same entities or agents as another agent. Here, we take a relatively simple approach to incentivizing joint attention in a multi-agent system. At each timestep, we produce a single $h \times w$ attention map for each agent $k$, by taking the mean value over each of the $m$ attention heads. Let $A^k_t$ be the mean attention map for agent $k$ at timestep $t$. We then measure the difference between $A^k_t$ and every other agent's attention map using the Jenson-Shannon Divergence (JSD):
\begin{align}
    r^{JA}_t &= - \sum_{j=1}^K\sum_{k=1}^K JSD(A^k_t||A^j_t) \\
             &= - \sum_{j=1}^K\sum_{k=1}^K \frac{1}{2}D_{KL}(A^k_t||M^{jk}_t) + \frac{1}{2}D_{KL}(A^j_t||M^{jk}_t)
    \label{eq:jsd}
\end{align}
where
\begin{align}
    D_{KL}(A^k_t||A^j_t) &= \sum_{x,y} a^k_{x,y} \log \frac{a^k_{x,y}}{a^j_{x,y}} 
    \label{eq:kl}
    \\
    M^{jk}_t &= \frac{1}{2}(A^j_t + A^k_t)
\end{align}
Note that because this measure of joint attention is fully symmetric, it only needs to be computed once for all agents. We compare using JSD to KL in Appendix Section \ref{sec:appendix_results} Figure \ref{fig:divergences} and find that both objectives give similar performance, so we focus on JSD for the remainder of the paper.

The computed attention difference, $r^{JA}_t$ is added to each agent's reward for timestep $t$. Therefore,  each agent attempts to maximize their own environment reward $r^k_t$, and the shared joint attention intrinsic motivation, $r_t^{JA}$. Thus, agent $k$'s objective is:
\begin{align}
    J(\pi^k) = \mathbb{E}_\pi \Big[ \sum_{t=0}^{\infty} \gamma^t \, (r^k_{t+1} + \beta r^{JA}_t)\, | \, s_0 \Big]
\end{align}
where $\beta$ controls the importance of the joint attention objective. Because other agents' attention is likely to be noisy at the beginning of training, we initially begin training with $\beta=0$, and scale it up over a curriculum of many episodes. 

Note that by adding the joint attention objective as an additional reward, rather than directly minimizing JSD at each time step, we allow agents to trade-off a short term loss in joint attention for a long term gain in environment reward, or vice versa. In other words, we do not force agents to constantly match attention at every timestep, but instead allow them to optimize both environment reward and joint attention over the course of the episode. 

We believe this approach has several key advantages over prior MARL approaches, which we outline below.

\subsection{No centralized critic}
The combinatorial nature of the joint action space in MARL means that accurately learning a joint value function requires an exponential increase in sample complexity. Many existing approaches to multi-agent training use a centralized critic \cite{foerster2017counterfactual,lowe2017multi,gupta2017cooperative}, which is trained to estimate a joint value function given the actions of all agents. This approach becomes inaccurate and impractical as the number of agents increases. 

In contrast, our approach does not depend on a centralized critic, and therefore does not require experiencing many examples of all possible combination of actions that agents can take in order to accurately estimate the value of each combination. Rather, we use other agents' attention as a way of guiding agents to learn to attend to salient and relevant aspects of the environment. As the number of agents increases, computing which elements are important in the environment is likely to become \textit{more} accurate, since more agents contribute to the computation, and the attention incentive in Eq. \ref{eq:jsd} becomes less noisy. 

That said, in environments requiring coordination, exploration can still be expensive if agents fail to try the right actions at the same time. However, we hypothesize that the joint attention incentive will reduce the cost of exploration by guiding agents to try coordinated actions simultaneously. Take the game Stag Hunt as an example. In the spatially and temporally extended version of this game \cite{peysakhovich2017prosocial,leibo2017multi}, two agents must simultaneously step on a Stag which is moving around the environment. Although it is relatively unlikely that this event occurs through random exploration, if both agents are simultaneously attending to the Stag, they may be more likely to take actions related to it at the same time. 

\subsection{Computational complexity}
The insight that incentivizing coordination could reduce the cost of multi-agent exploration was explored in \cite{jaques2018intrinsic}, which gave agents an intrinsic motivation to increase the causal influence, or mutual information, between their action and the actions of other agents. While this approach did not require a centralized controller, it was extremely computationally expensive. To compute the influence reward for each of $K$ independent agents required sampling $|\mathcal{A}|$ actions and computing predictions about how each of the other agents' predicted action would change. 
Computing this influence reward required $O(|\mathcal{A}|K)$ additional network passes per timestep during training. 

Instead, our approach does not require passing any additional data through agents' networks. Each agent computes its attention weights while it is acting in the environment, and these are stored into the replay buffer. We can then compute the joint attention incentive in Eq. \ref{eq:jsd} without ever re-running the agent's network, leading to a simpler and more efficient approach. 

\section{Experimental Setup}
In this section, we describe the multiagent environments, baselines and ablations used to test the effectiveness of the joint attention incentive. For the multi-agent environments, we not only compare agents' performance over the course of training, but also how well agents can generalize to modified versions of the environments at test time. All code used in our experiments is available in open-source at \url{https://github.com/google-research/google-research/tree/master/social_rl/multiagent_tfagents/joint_attention}. Further training hyperparamters and evaluation procedures are outlined in Appendix \ref{sec:training}.

\subsection{Baselines and ablations}
We compare our algorithm against \textit{MADDPG} \cite{lowe2017multi}, a popular CTDE method with a centralized critic. A recent benchmarking paper found that for the type of coordinated navigation task studied in this paper, MADDPG provides superior performance to more computationally expensive methods, such as QMIX \cite{rashid2018qmix}, making MADDPG a strong baseline. We use the authors' implementation of MADDPG, from: \url{https://github.com/openai/maddpg}.

We also compare against two independent PPO architectures, each with fully decentralized training. The \textit{Independent PPO} agents use a simplified architecture which processes the input with convolutional layers and updates a recurrent LSTM policy; it does not use attention. The \textit{Attention} agents use the attention architecture described in Figure \ref{fig:attention_architecture}, without applying the joint attention bonus (and with fully decentralized training). This ablation gives an indication of how much the improved performance obtained with the joint attention agents can be attributed to the architecture.


\subsection{Environments}
We evaluate our algorithm on four multi-agent environments explained in detail in Appendix Section \ref{sec:envs} and Figure \ref{fig:envs}. 
Environments are randomly generated each episode, and are implemented in a multi-agent, fully-observed version of Minigrid \cite{gym_minigrid}. We are releasing them in open-source at $<$URL redacted$>$.

These first three environments require agents to coordinate with other agents in order to effectively solve the task. 
\textbf{Meetup} is a discrete version of the particle-world Meetup proposed in the MADDPG paper \cite{lowe2017multi}.
In Meetup, agents must collectively choose one of $K$ landmarks and congregate near it; the goal landmark changes depending on the current position of all agents.
\textbf{ColorGather} is a modified version of the coin game in \cite{raileanu2018modeling}, where agents must collect the same coin that other agents are currently collecting. 
\textbf{StagHunt} is a temporally and spatially extended version of the classic matrix game, which has also been investigated in several other papers (e.g. \cite{peysakhovich2017prosocial,nica2017learning,leibo2017multi}). The environment contains two tasks. The agents can either collect berries, worth one point for the agent that collects it, or hunt stags, which are worth 5 points for both agents. Hunting stags requires collaboration between the agents, as one agent must stand adjacent to the stag while the other agent collects it. 

\textbf{TaskList} is a task that can be solved by a single agent, but which comprises a hard exploration problem. Agents must learn to complete a series of tasks independently, in the correct order: pick up a key, open a door, pick up a ball, open a box, drop the ball, reach the goal. We use TaskList to test whether joint attention can help novice agents learn a task more quickly, by training novices to attend to the same elements of the environment as experts, even when the task does not explicitly require coordinating with the expert.


\begin{figure*}
\centering
\begin{subfigure}{.3\textwidth}
  \centering
  \includegraphics[width=\linewidth]{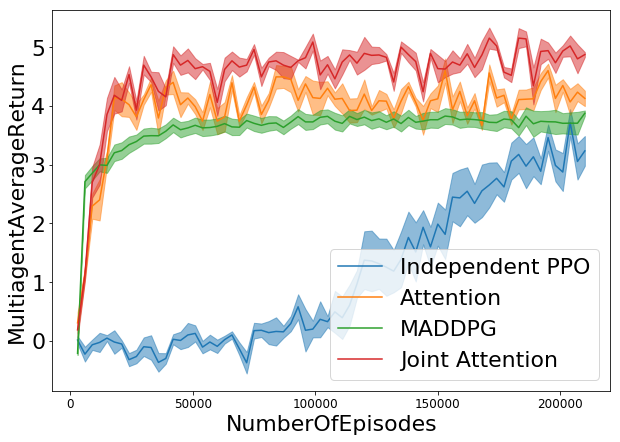}
  \caption{Meetup}
\end{subfigure}\hfil%
\begin{subfigure}{.3\textwidth}
  \centering
  \includegraphics[width=\linewidth]{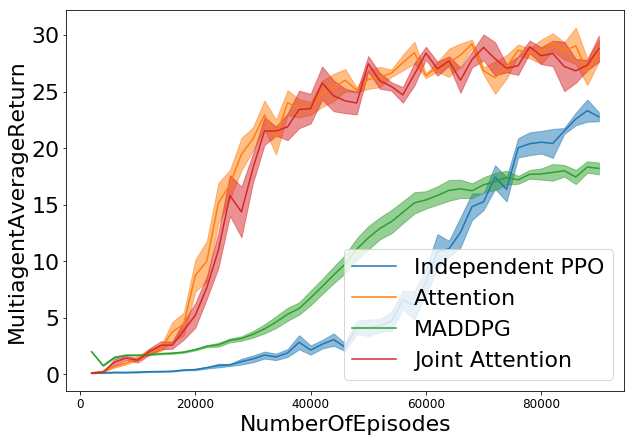}
  \caption{ColorGather}
\end{subfigure}
\begin{subfigure}{.3\textwidth}
  \centering
  \includegraphics[width=\linewidth]{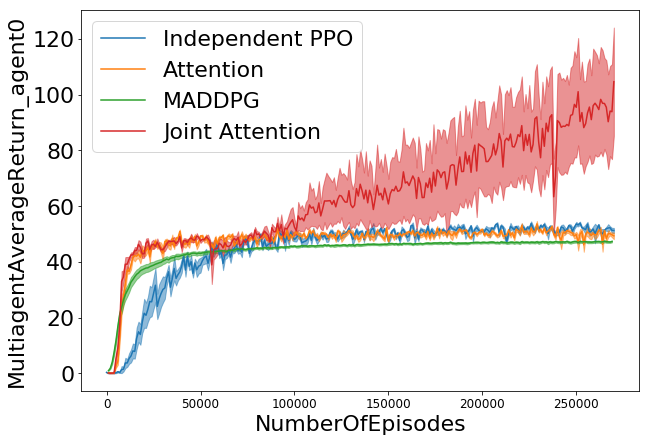}
  \caption{StagHunt}
\end{subfigure}
\caption{Collective average reward in multi-agent coordination environments. The independent PPO baseline learns slowly and reaches lower final performance on all three tasks. Simply adding the proposed attention architecture improves agents performance, enabling them to achieve higher scores than vanilla PPO and MADDPG, even though the agents are trained independently. MADDPG learns more quickly and stably than independent PPO, but converges to a suboptimal policy. Finally, the joint attention incentive provides clear benefits for increasing learning speed and final performance.}
\label{fig:results}
\end{figure*}

\section{Results}
This section investigates the following questions: i) Can joint attention enhance multi-agent coordination, and if so, how? (Sec. \ref{sec:results_coordination}); ii) What is its effect on generalization to new environments at test time? (Sec \ref{sec:results_generalization}; 
and iii) Can joint attention improve the ability of agents to learn from experts in their environment? (Sec \ref{sec:results_experts}).

\subsection{Multi-agent coordination}
\label{sec:results_coordination}
Figure \ref{fig:results} shows evaluation results throughout training for each technique in the multi-agent coordination environments. The Independent PPO baseline performs poorly, learning more slowly than the other techniques. This is likely due to the exponential sample complexity inherent in multi-agent exploration. 
However, simply using the proposed attention architecture leads to competitive performance, even when agents are trained independently. In fact, the performance of the architecture is superior to that of MADDPG, in spite of the fact that MADDPG uses a centralized critic, and the attention architecture agents were trained in a fully decentralized fashion. This is consistent with recent findings on multi-agent StarCraft \cite{schroeder2020independent}, which showed that independent PPO training can surpass CTDE MARL algorithms given a sufficiently advanced PPO implementation. 

\begin{figure*}
\centering
\begin{subfigure}{.5\textwidth}
  \centering
  \includegraphics[width=\linewidth]{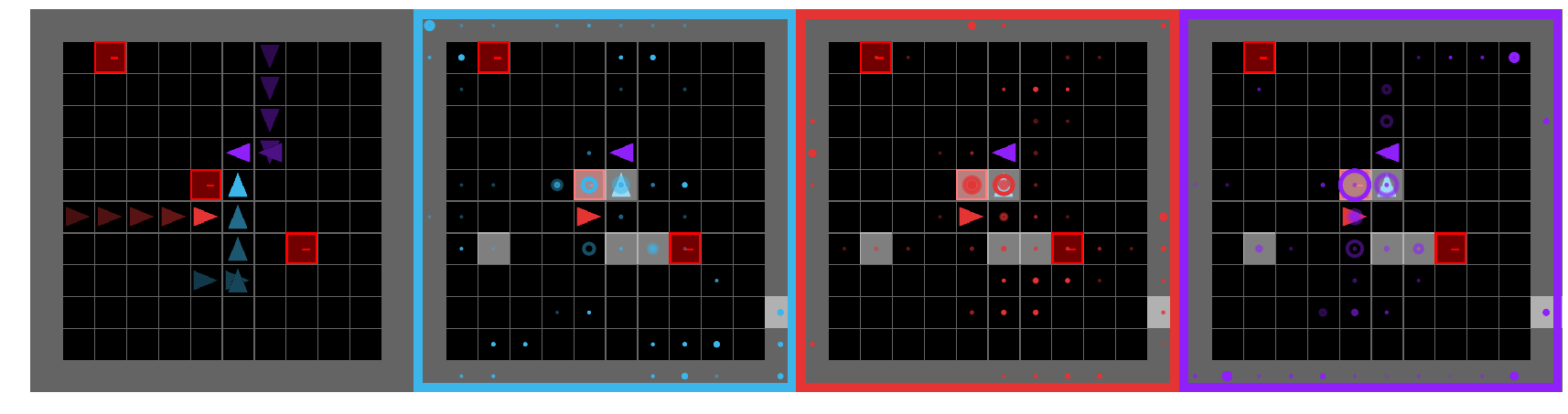}
  \caption{Meetup without joint attention}
  \label{fig:a_meetup}
\end{subfigure}\hfil%
\begin{subfigure}{.5\textwidth}
  \centering
  \includegraphics[width=\linewidth]{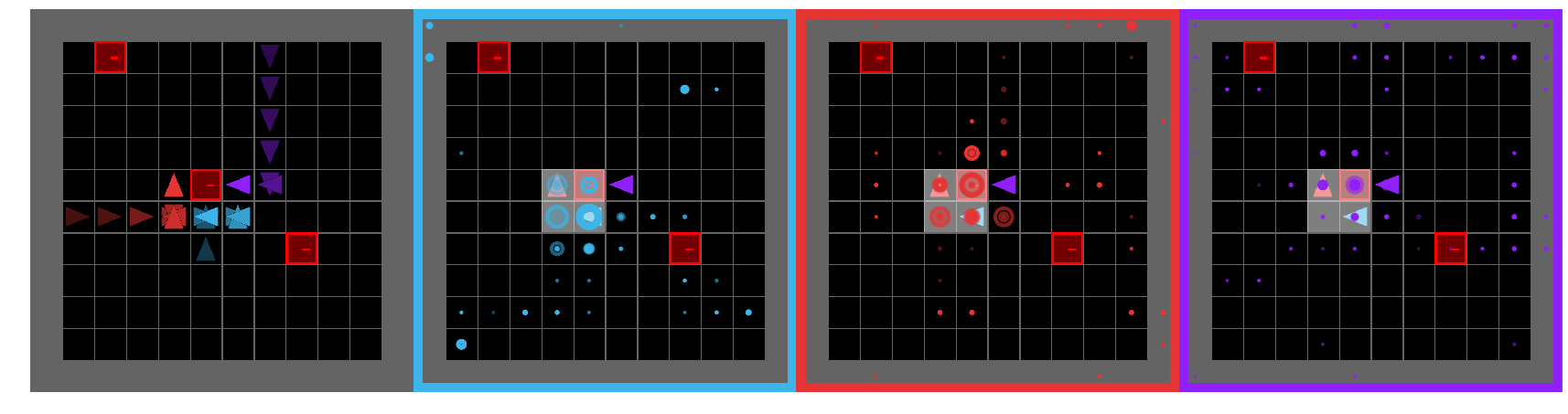}
  \caption{Meetup with joint attention incentive}
  \label{fig:ja_meetup}
\end{subfigure}
\begin{subfigure}{.5\textwidth}
  \centering
  \includegraphics[width=\linewidth]{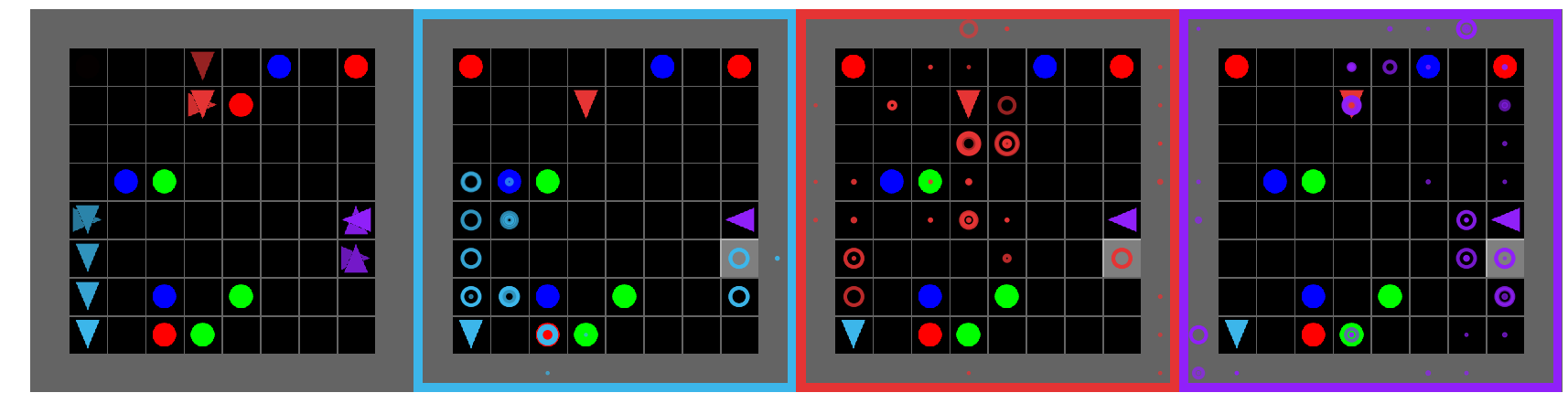}
  \caption{ColorGather without joint attention}
  \label{fig:a_colorgather}
\end{subfigure}\hfil%
\begin{subfigure}{.5\textwidth}
  \centering
  \includegraphics[width=\linewidth]{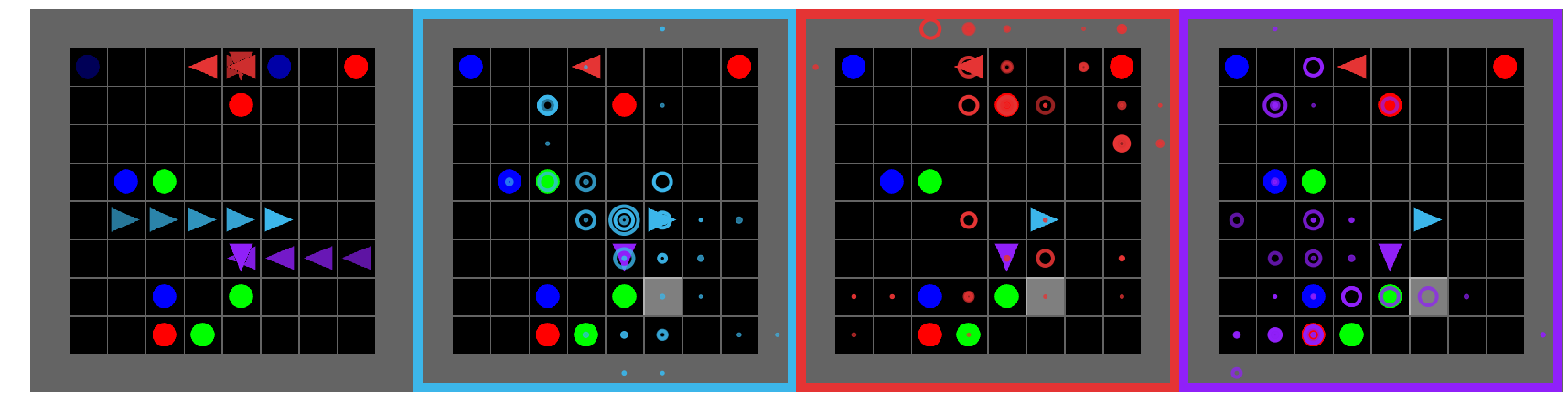}
  \caption{ColorGather with joint attention incentive}
  \label{fig:ja_colorgather}
\end{subfigure}
\begin{subfigure}{.5\textwidth}
  \centering
  \includegraphics[width=\linewidth]{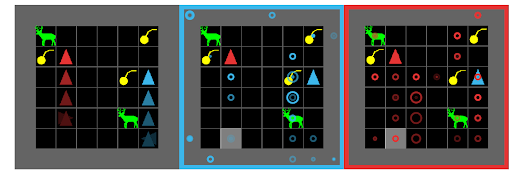}
  \caption{StagHunt without joint attention}
  \label{fig:a_staghunt}
\end{subfigure}\hfil%
\begin{subfigure}{.5\textwidth}
  \centering
  \includegraphics[width=\linewidth]{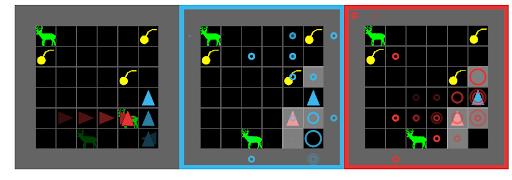}
  \caption{StagHunt with joint attention incentive}
  \label{fig:ja_staghunt}
\end{subfigure}
\caption{Comparison of learned attention maps. Within each figure, the left-most square is the ground-truth state of the environment, with the position of agents and objects over the past five timesteps shown using faded versions of the element. The next images show the attention of the individual agents, outlined in the colour of the agent it represents. The circles are the elements the agents are attending to, where size represents the attention strength (larger is stronger), and the transparency represents the timestep (faded is farther in the past). Squares highlighted in grey represent elements that are mutually attended to by agents at the same time. The left column shows that agents have learned to use attention to focus on salient elements of the environment. The right column demonstrates that adding the joint attention incentive helps agents coordinate their attention on the same elements, enabling them to perform coordinated behaviors (e.g. catching the stag in (f)).}
\label{fig:attention_map}
\end{figure*}

Adding the joint attention incentive significantly speeds learning and improves performance above MADDPG in all three environments. In ColorGather, joint attention performs equivalently to the attention architecture. Since there is no penalty for picking up the wrong object, agents can effectively solve ColorGather by focusing on the objects closest to themselves, rather than the objects attended to by other agents (and indeed, we found this happens about 25\% of the time with both types of agents). 
However, in Meetup and StagHunt, joint attention significantly improves agents' ability to coordinate. We note that in StagHunt, all baseline methods only collect berries, and do not learn the difficult coordination task of hunting the stag together. In contrast, the joint attention agents are able to mutually attend to the stag at the same time (as shown in Figure \ref{fig:attention_map}), enabling them to learn to successfully complete the task.

Figure \ref{fig:attention_map} shows sample attention maps for agents trained with and without joint attention. Even without the joint attention incentive, the agents learn to focus on salient aspects of the environment, such as multiple meetup points (\ref{fig:a_meetup}), or berries (\ref{fig:a_staghunt}). 
In ColorGather, agents attend to the region around themselves, whether they are given the joint attention incentive (\ref{fig:ja_colorgather}) or not (\ref{fig:a_colorgather}), which is consistent with the results of Figure \ref{fig:results}.
However, when the joint attention bonus is applied to Meetup and StagHunt, agents are able to coordinate their attention to simultaneously focus on the same elements. In Figure \ref{fig:ja_meetup}, all agents' attention is heavily focused on each other, and the single waypoint where they are meeting. In Figure \ref{fig:ja_staghunt}, agents' attention is focused on the stag and the area around it, which enables them to catch the stag. 
These results help explain how agents can use shared attention to make the challenging problem of joint exploration easier.

\subsection{Generalization}
\label{sec:results_generalization}
\begin{figure*}
\centering
\begin{subfigure}{.31\textwidth}
  \centering
  \includegraphics[width=\linewidth]{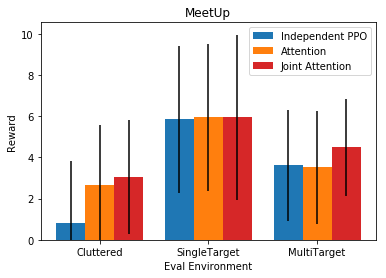}
  \caption{Meetup}
\end{subfigure}\hfil%
\begin{subfigure}{.31\textwidth}
  \centering
  \includegraphics[width=\linewidth]{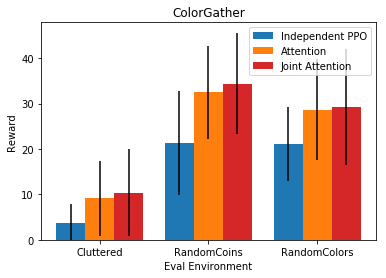}
  \caption{ColorGather}
\end{subfigure}
\begin{subfigure}{.31\textwidth}
  \centering
  \includegraphics[width=\linewidth]{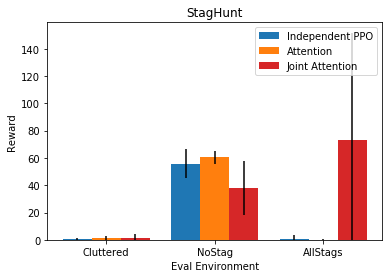}
  \caption{StagHunt}
\end{subfigure}
\caption{Results of generalization of the best-performing seed to perturbations of the environment at test time. The attention architecture and joint attention incentive lead to equivalent or improved generalization.}
\label{fig:generalization}
\end{figure*}
We are interested in assessing the generalization of our algorithm, because deep RL techniques often fail to generalize to even slight modifications of the training environment \cite{cobbe2019quantifying,farebrother2018generalization,packer2018assessing}. To that end, we introduce modifications to the environments at test time, and measure zero-shot performance under these modifications. For example, we add new obstacles, change the number of meetup points or coin colors, or remove all stags. A full description of the test environments is available in Appendix Section \ref{sec:gen_envs} and Figure \ref{fig:gen_envs}. 
On each environment, we evaluate our best performing policy for each method for 30 episodes. 

Figure \ref{fig:generalization} shows the zero-shot performance of each method in the modified test environments. 
With the exception of the NoStag environment, the attention and joint attention agents always generalize equally well or better than the baselines. The reduction in joint attention performance in the NoStag environment is because the baseline learned to ignore stags and only pick up berries (as seen in Figure \ref{fig:a_staghunt}). 
Similarly, the large improvement in the AllStags environment is due to the fact that only the joint attention agents learned to catch stags. 
Overall, these results help establish that the joint attention architecture and incentive do not decrease agents' ability to generalize to new environments, and may improve it by filtering out irrelevant elements like walls.

\subsection{Learning from experts}
\label{sec:results_experts}
\begin{figure}
\centering
\includegraphics[width=.8\linewidth]{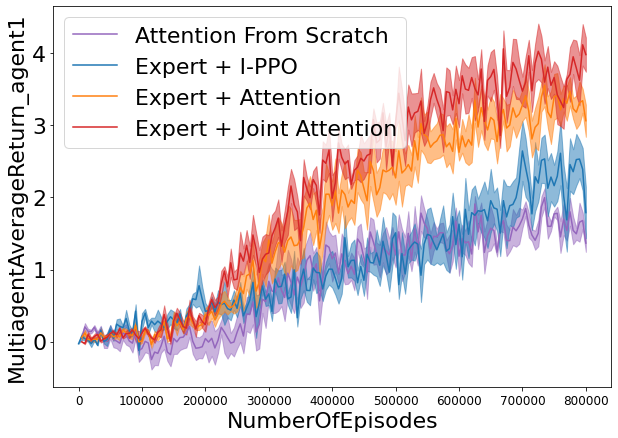}
\caption{Results of learning jointly with an expert vs. learning alone in the TaskList environment. The joint attention incentive enables agents to learn more quickly from experts, even in a hard exploration task that does not explicitly require coordinating with another agent.}
\label{fig:experts}
\end{figure}
We hypothesize that when there are expert agents present in the environment, joint attention could scaffold the learning of novice agents, by training them to focus on the parts of the environment relevant for solving the task. Therefore, we take expert agents pre-trained in TaskList, and place those experts in the same environment as newly initialized, novice agents. Since joint attention enhances the ability of human caregivers to teach their children \cite{tomasello1995joint,mundy1990longitudinal}, we expect it could provide similar benefits to agents in a multi-agent system. However, prior work \cite{ndousse2020social} has shown that vanilla model-free RL agents struggle to learn from experts in their environment.

Figure \ref{fig:experts} shows the training curves resulting from placing two novice agents into the TaskList environment with a single pre-trained expert. 
Unlike in prior work \cite{ndousse2020social}, both the attention architecture and joint attention incentive show improved learning in the presence of experts as compared to learning from scratch. The joint attention incentive provides a further increase to learning speed above the architecture, reducing the sample complexity of reaching optimal performance. We hypothesize this is because joint attention provides an additional signal which guides the novice agent's focus. These results demonstrate that joint attention can help in learning from experts even in tasks where the goal is not about coordination.

\section{Discussion}
Overall, our results suggest that joint attention can enhance social learning, generalization, and coordination with other agents. We hypothesize that joint attention reduces the cost of exploring over the joint action space. This is because when agents are focused on the same elements of the environment at the same time (as in Figure \ref{fig:ja_staghunt}), they are more likely to try actions related to that element simultaneously. Therefore, joint attention provides a relatively simple and efficient way to reduce the cost of multi-agent exploration. Further, joint attention enables agents to learn from each other's experience by guiding their attention to the important parts of the environment. Finally, we show that our method provides equal or improved generalization performance, which is an important consideration for deep RL algorithms. These results provide promising evidence that joint attention may be a useful inductive bias for RL agents.

\textbf{Limitations and future work.}
We adopted the centralized training, decentralized execution (CTDE) framework popular in prior multi-agent deep RL work (e.g. \cite{lowe2017multi,foerster2017counterfactual,rashid2018qmix,sunehag2018value}). However, CTDE assumes that all agents are trained together under the control of a centralized entity, which does not apply for many real-world scenarios which involve coordinating with humans, such as autonomous driving or household robotics. In future, we would like to improve upon techniques for inferring human attention (e.g. \cite{hoffman2006probabilistic,kozima2001robot}), and train agents to maintain joint attention with human interlocutors. We hypothesize this could facilitate human-AI coordination.

In this paper we consider a uniform joint attention reward which is the same for all agents. Interesting future directions would be to reward as for matching a median attention map (to avoid penalties for not matching with errant outlier agents), or rewarding how well attention is matched with a single other agent that is closest in attention space.

\textbf{Acknowledgements.}
We would like the thank Sergey Levine, Jie Tan, and the TFAgents team, especially Oscar Ramirez and Summer Yue for the helpful discussions that contributed to this work.



\bibliographystyle{icml2021}  
\bibliography{jointattention}  
\vfill

\clearpage

\begin{figure*}[ht!]
\centering
\begin{subfigure}{.24\textwidth}
  \centering
  \includegraphics[width=\linewidth]{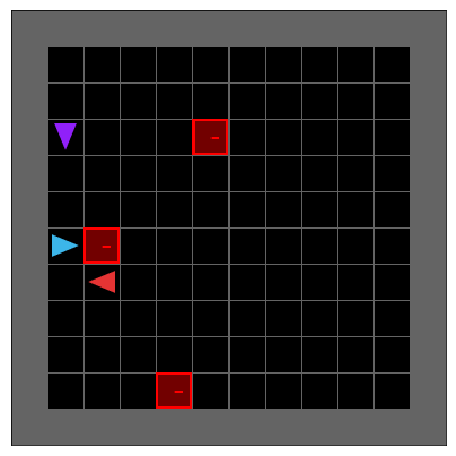}
  \caption{Meetup}
  \label{fig:meetup}
\end{subfigure}\hfil%
\begin{subfigure}{.24\textwidth}
  \centering
  \includegraphics[width=\linewidth]{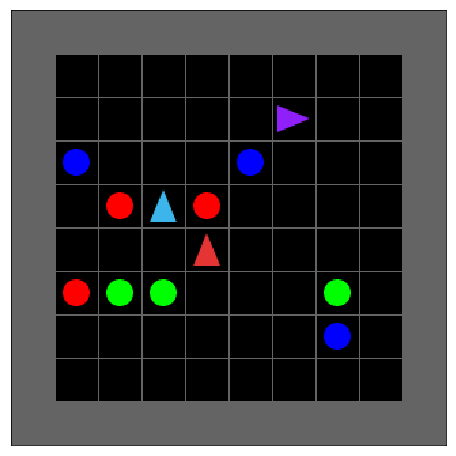}
  \caption{ColorGather}
  \label{fig:colorgather}
\end{subfigure}\hfil%
\begin{subfigure}{.24\textwidth}
  \centering
  \includegraphics[width=\linewidth]{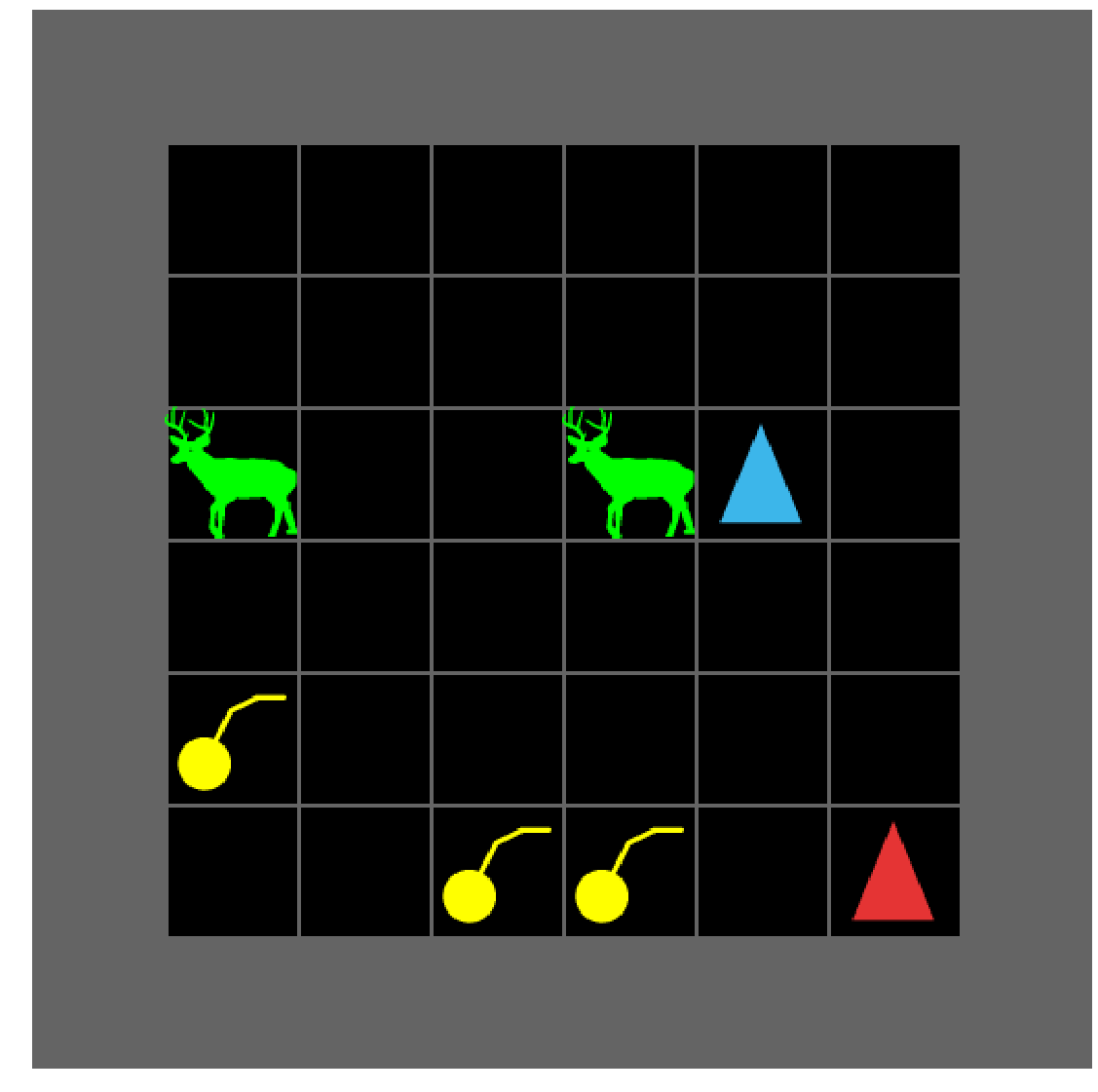}
  \caption{StagHunt}
  \label{fig:staghun}
\end{subfigure}\hfil%
\begin{subfigure}{.24\textwidth}
  \centering
  \includegraphics[width=\linewidth]{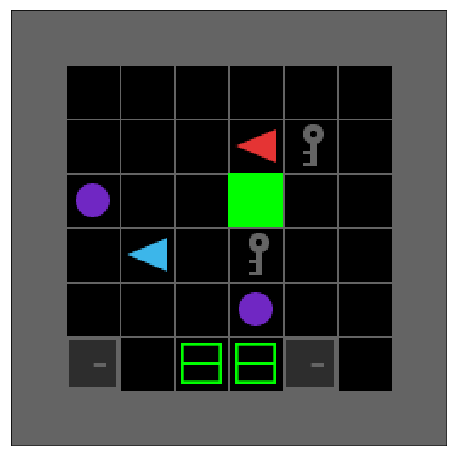}
  \caption{TaskList}
  \label{fig:doorkey}
\end{subfigure}\hfil%
\caption{Multi-agent environments. (a-c) are multi-agent environments in which agents (triangles) must coordinate effectively with each other to solve the task. In Meetup (a), agents are rewarded for congregating at the same landmark (red box) as other agents, which updates dynamically as the location of other agents changes. Similarly, the color of objects that agents must pick up in ColorGather (b), changes based on which object other agents have collected most. In StagHunt (c), agents can receive individual rewards for collecting berries, but will receive a higher reward if they cooperate to step on the stag while the other agent is adjacent to it. The last environment, TaskList (d), is used for testing whether joint attention helps novice agents learn from an expert in the same environment, but does not require coordination. Instead, agents must learn to independently complete a complex series of tasks such as picking up objects and using keys to open doors. We use TaskList to asses if agents can learn effectively from viewing experts completing the task.
}
\label{fig:envs}
\end{figure*}

\appendix

\section{Neural network architecture details}
\label{sec:training}
We use the same network architecture across all agents. The environment observation is first processed by a 3x3 convolutional layer with 64 filters, stride 1, padding to maintain the size of the input, and ReLU activations. In our attention architecture, we concatenate a spatial basis of depth 8 to the image features, then use 4 attention heads of depth 16 to create K and V. The scalar inputs are processed with a single fully connected layer of size 5. For the policy network we use an LSTM with cell size 64, followed by two fully connected layers with hidden size 64. The value network is identical to the policy network and shares no weights. During training we anneal  the weight of the bonus reward $\beta$ linearly from 0 to $10^{-2}$ over the first 200000 steps. We optimize the policy and value networks using Adam with a learning rate of $10^{-4}$. 

\subsection{Spatial basis equations}
\label{sec:spatial_basis}
Our spatial basis matrix $S$ is a 2D version of the position encoding (PE) found in \cite{vaswani2017attention}, where the first $c_s / 2$ basis vectors are the position encoding of the horizontal coordinate $x$, and the second $c_s / 2$ are the position encoding of the vertical coordinate $y$. A similar 2D position encoding is proposed in \cite{wang2019translating}.
The following equations are used to produce each $h \times w$ layer of the spatial basis matrix $S$, where $i$ varies from $1$ to $c_s / 4$. 
\begin{align*}
PE(x, y, 2i) &= \sin(x / 100^{4i/c_s}) \\
PE(x, y, 2i + 1) &= \cos(x / 100^{4i/c_s}) \\
PE(x, y, 2i + D / 2) &= \sin(y / 100^{4i/c_s}) \\
PE(x, y, 2i + D / 2 + 1) &= \cos(y / 100^{4i/c_s}) \\
\end{align*}

\section{Training and hyperparameters}
\label{sec:hp}
To find our hyperparameter settings, we performed a grid search across the following: learning rate $\in [10^{-3}, 10^{-1}]$, batch size $\in [32, 128]$, training epochs per batch $\in [3, 10]$, number of convolutional layers $\in [1, 3]$, number of fully connected layers $\in [1, 3]$, fully connected layer size and LSTM layer size $\in [32, 256]$, attention bonus weight $\in [10^{-4}, 10^{-1}$, attention bonus scale up steps $\in [10^3, 10^6]$.

For MADDPG, we use the default training hyperparameters from the open sourced code. The replay buffer consists of the last 1024 episodes, from which we sample a batch of 1024 steps for training. The networks are trained with Adam with learning rate $10^{-2}$ for one gradient step every 100 environment steps, and updated with polyak averaging with $\tau = 0.99$. The architecture used is exactly identical to the I-PPO baseline, without the recurrent component. The primary reason we incorporate a recurrent network in our architecture is to provide context to the attention heads. As the environments are fully observed, the recurrent component is not necessary for solving the task. 

To evaluate our method and baselines, we pause the training every 3000 episodes and evaluate for 10 episodes using greedy decentralized execution using the TFAgents framework \cite{TFAgents}. For each curve, we plot the mean and one standard deviation range across 5 sets of agents. Each training and evaluation run takes between 1 and 3 days, depending on environment, on a single cloud based CPU instance, such as those found on AWS or Google Cloud.

\section{Environment details}
\label{sec:envs}
The environments studied in this paper are visualized in Figure \ref{fig:envs}.
In each, agents receive a fully observed encoding of the environment, where object type (agent, ball, stag, wall, door, etc) and color are identified, as well as the agent's own position and the direction it is facing. Rather than using pixels, our agents see an encoded version of the environment with three channels, as proposed in \cite{gym_minigrid}.

\textbf{Meetup} is a discrete version of the particle-world Meetup proposed in the MADDPG paper \cite{lowe2017multi}.
In Meetup, agents must collectively choose one of $K$ landmarks and congregate near it. At each timestep, each agent receives reward equal to the change in distance between itself and the landmark closest to all three agents. The goal landmark changes depending on the current position of all agents. When all $K$ agents are adjacent to the same landmark, the agents receive a bonus of 1 and the episode ends. 

\textbf{ColorGather} is a modified version of the coin game in \cite{raileanu2018modeling}. 
In ColorGather, coins of N different colors are randomly placed around the environment. The agents must collectively choose one color and collect coins of that color. Each time a coin is collected, a new coin of that color is placed. All agents are rewarded for the first coin collected regardless of color, and then for each coin that matches the most-collected color that episode. 

\textbf{StagHunt} is a temporally and spatially extended version of the classic matrix game, which has also been investigated in several other papers (e.g. \cite{peysakhovich2017prosocial,nica2017learning,leibo2017multi}). The environment contains two tasks. The agents can either collect berries, worth one point for the agent that collects it, or hunt stags, which are worth 5 points for both agents. Hunting stags requires collaboration between the agents, as one agent must stand adjacent to the stag while the other agent collects it. 

\subsection{Generalization environments}
\label{sec:gen_envs}
\begin{figure*}
\centering
  \includegraphics[width=.8\linewidth]{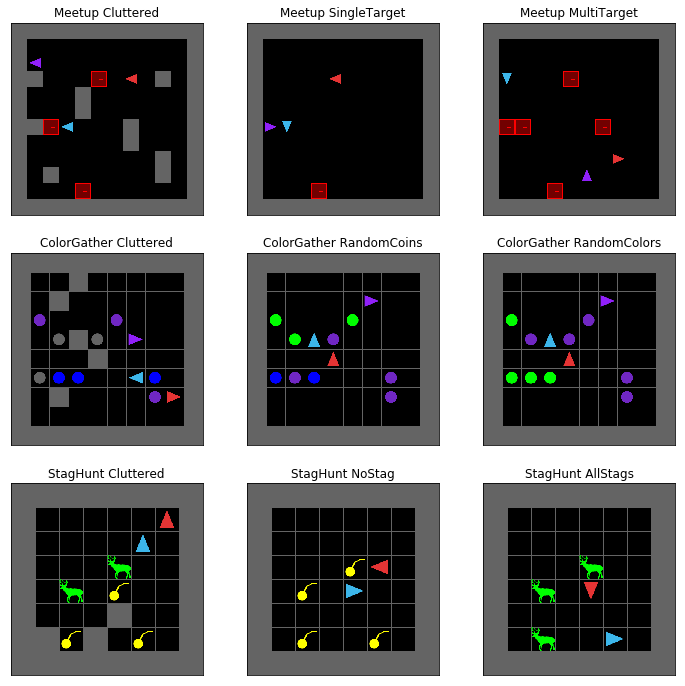}
\caption{Visualization of environments used to test how well agents can perform on tasks that are outside the distribution of training tasks. Results are shown in \ref{sec:results_generalization}}
\label{fig:gen_envs}
\end{figure*}

Figure \ref{fig:gen_envs} shows each modified test environment designed to assess how well agents can generalize to tasks outside the training distribution.
In the `Cluttered' environment, 10\% of the open space is filled with walls, which constitute a new object not encountered during training. The `SingleTarget' environment has only a single meetup point, while the `MultiTarget' environment has 5, rather than 3. In the `RandomCoins' environment, the number of coins of each color varies between 1 and 4 each episode, rather than always being 3, while the `RandomColors' environment has a random number of colors between 2 and 4, rather than always being 3. The `NoStag' environment has only berries and no stags, while the `AllStags' environment has only stags and no berries. Note, that some of the environments are easier, and some are more difficult. For example, `NoStag' environment is easier than `StagHunt', while `AllStags' is more difficult. 
 and which are described in Section \ref{sec:results_generalization}.

\section{Additional results}
\label{sec:appendix_results}
\begin{figure}
\vspace{-0.15in}
\centering
\includegraphics[width=.8\linewidth]{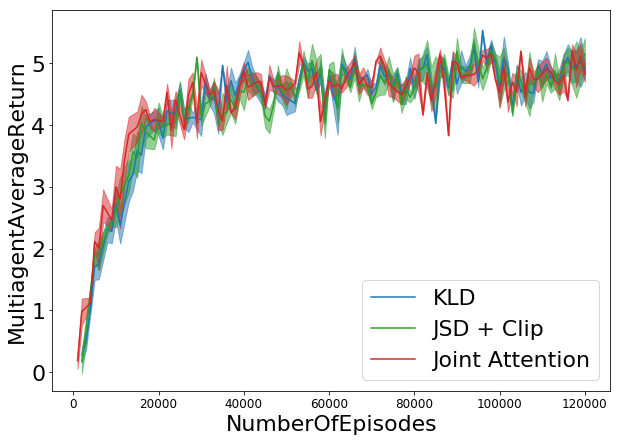}
\vspace{-0.1in}
\caption{Comparison of different methods for calculating the divergence between agents' attention for the joint attention incentive, in Meetup. There are no significant differences in performance between KL, JSD, and clipped JSD, suggesting the method is robust to the choice of divergence metric. Results for the other environments were similar. }
\label{fig:divergences}
\vspace{-0.15in}
\end{figure}

\subsection{Comparing joint attention metrics}
\label{sec:results_metrics}
In addition to JSD, we evaluate two other methods for computing the attention bonus. Kullback-Liebler (KL) divergence (Eq. \ref{eq:kl}) is a natural choice for computing the shared information between the attention weights of two agents. We also evaluate a clipped version of JSD, where prior to normalization all logits less than a threshold are clipped to negative infinity for the bonus computation only. The actual weights used in the attention computation remain unchanged. We hypothesized that clipping would improve the bonus by making it less noisy, since it would not be affected by small differences in regions of the attention map that were not agents' focus.

Figure \ref{fig:divergences} shows the performance of the different methods used for computing the joint attention incentive. Both JSD and KL provide similar performance, showing the method is robust to the choice of divergence metric. JSD may still be preferred over KL for its simplicity, since JSD produces a single $r^{JA}$ objective that is shared by all agents. We also see that clipping the JSD objective is unnecessary.

\end{document}